# Morphological Analysis of Japanese Hiragana Sentences using the Bi-LSTM CRF Model

Jun Izutsu[1] and Kanako Komiya[2]

[1]Ibaraki University, Japan
[2]Tokyo University of Agriculture and Technology, Japan

## Abstract

*This study proposes a method to develop neural models of the morphological analyzer for Japanese Hiragana sentences using the Bi-LSTM CRF model. Morphological analysis is a technique that divides text data into words and assigns information such as parts of speech. This technique plays an essential role in downstream applications in Japanese natural language processing systems because the Japanese language does not have word delimiters between words. Hiragana is a type of Japanese phonogramic characters, which is used for texts for children or people who cannot read Chinese characters. Morphological analysis of Hiragana sentences is more difficult than that of ordinary Japanese sentences because there is less information for dividing. For morphological analysis of Hiragana sentences, we demonstrated the effectiveness of fine-tuning using a model based on ordinary Japanese text and examined the influence of training data on texts of various genres.*

## Keywords

*Morphological analysis, Hiragana texts, Bi-LSTM CRF model, Fine-tuning, Domain adaptation.*

## 1. Introduction

### 1.1. Components of the Japanese Language and the Acquisition Process

Japanese sentences contain various kinds of character, such as Kanji (Chinese character), Hiragana, Katakana, numbers, and alphabet, making it difficult to learn. Japanese speakers usually learn Hiragana first in their school days because the number of characters is much smaller than the Kanji; Hiragana has 46 characters, and Japanese use thousands of Kanji. Most Japanese sentences are composed of all kinds of characters and are called Kanji-Kana mixed sentences.

However, it is difficult for many non-Japanese speakers to learn thousands of Kanji, so children and new Japanese language learners use Hiragana.

### 1.2. Morphological Analysis of Hiragana Sentences

Morphological analysis is a technique that divides natural language text data into words and assign information such as parts of speech. In Japanese, morphological analysis is one of the core technologies for natural language processing because the Japanese language does not have word delimiters between words. Morphological analyzers like MeCab[1], Chasen and Juman++[2,3] are now commonly used for morphological analysis. However, since the above systems target Kanji-Kana mixed sentences, it is challenging to perform morphological analysis using only Hiragana sentences.

Morphological analysis of Hiragana-only sentences is more challenging than morphological analysis of Kanji-Kana mixed sentences. If the text is a mixture of Kanji and Kana (Hiragana and Katakana), it will be divisible between Kanji, Hiragana, and Katakana. However, if the text consists only of Hiragana, there will be less information for dividing. Therefore, we propose to fine-tune the model of Kanji-Kana mixed sentences and investigated whether the accuracy of morphological analysis of Hiragana sentences can be improved by inheriting the information to be divided into words.

The Bi-LSTM CRF model was used to develop a morphological analyzer for Hiragana sentences in this paper. We used two types of training data: Wikipedia and Yahoo! Answers in the Balanced Corpus of Contemporary Written Japanese, in our studies to investigate the influence of the genre of the training data. We also fine-tune both data to examine the effect of various genres on the text. [4]

The following are the four contributions of this paper.

- Developed a morphological analyzer for Hiragana sentences using the Bi-LSTM CRF model,
- Demonstrated the effectiveness of fine-tuning using a model based on Kanji-Kana mixed text,
- Examined the influence of training data of morphological analysis on texts of various genres, and
- Demonstrated the effectiveness of fine-tuning using data from Wikipedia and Yahoo! Answers.

In this paper, we report the results of these experiments.

## 2. RELATED WORK

Izutsu et al. (2020) [5] converted MeCab'sipadic dictionary into Hiragana and performed morphological analysis on Hiragana sentences using a corpus consisting only of Hiragana. To our knowledge, only this work developed a morphological analyzer for only Hiragana sentences.We also developed a morphological analyzer for only Hiragana sentence, but we try to achieve this goal without any dictionary.

There are some studies focus on morphological analyzers for Hiragana-highly-mixed sentences and most of them treated a lot of Hiragana words as noises or broken Japanese. For example, Kudo et al. (2012) [6] used generative model to model the process of generating Hiragana noise-mixed sentence. They proposed using a large-scale web corpus and EM algorithm to estimate the model's parameters to improve the analysis of Hiragana noise-mixed sentences. Osaki et al. (2016) [7] constructed a corpus for broken Japanese morphological analysis. They defined new parts of speech and used them for broken expressions. Fujita et al. (2014) [8] proposed an unsupervised domain adaptation technique that uses the existing dictionaries and labeled data to build a morphological analyzer by automatically transforming them for the features of the target domain. Hayashi and Yamamura (2017) [9] reported that adding Hiragana words to the dictionary can improve the accuracy of morphological analysis.

We used Bi-LSTM model to develop a morphological analyzer. Ma et al. (2018) [10] developed a word segmentation model for Chinese using the Bi-LSTM model.They reported that word segmentation accuracy achieved better accuracy on public datasets than the Bi-LSTM model, compared to models based on more complex neural network architectures.

Also, Thattianaphanich and Prom-on (2019) [11] developed the Bi-LSMT CRF model and performed named entity recognition extraction in Thai. In Thai, there are linguistic problems such as lack of linguistic resources and boundary indices between words, phrases, and sentences. Therefore, they prepared word representations and learned text sequences using Bi-LSTM and CRF to address these problems.

In work on Japanese morphological analyzers, Tolmachev et al. created a morphological model using neural networks and semi-supervised learning, and showed that their method performed comparably to traditional dictionary-based state-of-the-art methods, and could even outperform them when trained on a combination of human-annotated and automatically annotated data [12].

In addition, Chen et al. created an LSTM-based neural network model for Chinese word segmentation [13]. Although most of the current state-of-the-art methods for Chinese word segmentation are based on supervised learning, they report that their LSTM-based model outperforms traditional neural network models and state-of-the-art methods.

## 3. METHODS

In this research, we used the Bi-LSTM CRF model to generate a morphological analyzer for Hiragana sentences.We trained and compared the following five models for Hiragana sentences in our studies.

1. the Hiragana Wiki model
2. the Hiragana Yahoo! model
3. the Kanji-Kana Wiki+ Hiragana Wiki model
4. the Kanji-Kana Wiki+ Hiragana Wiki+ Hiragana Yahoo! model
5. the Hiragana Wiki+ Hiragana Yahoo! model

The generation processes of the five models for Hiragana sentences are shown in Figure 1.

### 3.1. Hiragana Wiki Model

The Hiragana Wiki model is the model generated in training B (Figure 1). We used the data from Wikipedia, where Kanji-Kana mixed sentences are converted to Hiragana, for training data. Hiragana is a phonogram, and Kanji could be converted into Hiragana according to its pronunciation. We used MeCab's reading data as a pseudo-correct answer for the conversion because Wikipedia does not have Hiragana-only data. Here, UniDic was used as a dictionary of MeCab. [14]

### 3.2. Hiragana Yahoo! Mode

The model generated in training D (Figure 1) is the Hiragana Yahoo! model. We used reading data from Yahoo! Answers as training data. We compared its accuracy to a Hiragana Wiki + Hiragana Yahoo! model (training F in Figure 1) and Kanji-Kana Wiki+ Hiragana Wiki+ Hiragana Yahoo! model (training E in Figure 1).

### 3.3. Kanji-Kana Wiki+ Hiragana Wiki Model

In this paper, we proposed fine-tuning using a model with Kanji-Kana mixed sentences. The Kanji-Kana Wiki+ Hiragana Wiki model is generated using the original Wikipedia data and Hiragana Wikipedia data, which are the actual data converted into only Hiragana. We fine-tuned Kanji-Kana Wiki model, the model trained by the original Wikipedia data (training A in Figure 1), with Hiragana Wikipedia data, which is Wikipedia data automatically converted into Hiragana sentences (training C in Figure 1). Kanji-Kana mixed sentences contain many clues for morphological analysis, such as borderline between Kanji and Hiragana and information about Kanji. Therefore, it is expected to improve accuracy. By comparing the Hiragana Wiki model and this model, we can assess the effectiveness of fine-tuning based on Kanji-Kana mixed Wikipedia data, when we only have Wikipedia data.

### 3.4. Kanji-Kana Wiki+ Hiragana Wiki+ Hiragana Yahoo! Model

Training E in Figure 1 generates the Kanji-Kana Wiki+ Hiragana Wiki+ Hiragana Yahoo! model. To improve accuracy, we used both Wikipedia and Yahoo! Answers as training data.

We fine-tuned the Kanji-Kana Wiki+ Hiragana Wiki model with Hiragana Yahoo! Answers data for this model.

### 3.5. Hiragana Wiki+ Hiragana Yahoo! Model

The Hiragana Wiki+ Hiragana Yahoo! model is the model generated in training F in Figure 1. As training data, we used Hiragana Wikipedia data and Hiragana Yahoo! Answers data. We fine-tuned the Hiragana Wiki model with Hiragana Yahoo! Answers.

We can see the effectiveness of the Kanji-Kana Wiki+ Hiragana Wiki model, i.e., how much the Kanji-Kana mixed model affects this model, when we have both Wikipedia data and Yahoo! Answers by comparing the accuracy of this model to Kanji-Kana Wiki+ Hiragana Wiki+ Hiragana Yahoo! model.

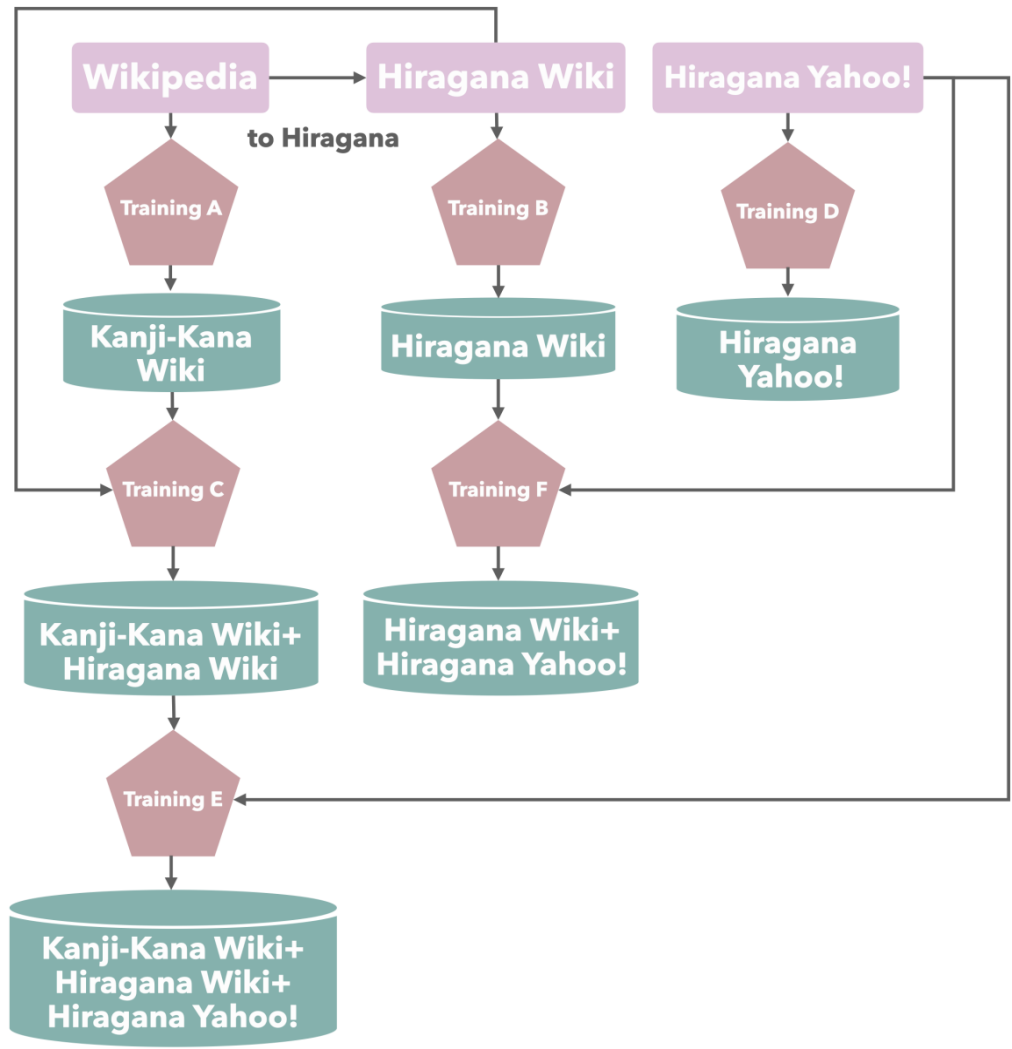

Figure 1. Model generation processes of the five models for Hiragana sentences

## 4. EXPERIMENT

### 4.1. Network Architecture

We developed Bi-LSTM CRF models using Bi-LSTM CRF module of Pytorch.

Table 1 summarizes the hyperparameters used in this experiment. The dimensions for the network architecture were determined according to the preliminary experiments.

Table 1. Hyperparameters used in the experiment

| EMBEDDING_DIM | 5 |
|---|---|
| HIDDEN_DIM | 4 |
| Lr | 0.01 |
| Weight decay | 1e-4 |
| Optimizer | sgd |
| Epoch number | 15 |

The tag size is 38 for all models and the vocabulary size varies according to the model because it is the total character type number of the training corpus. The vocabulary size of the models that use only Hiragana sentences are 192 and that of the model that uses Kanji-Kana Wiki data was 2,742.

### 4.2. Training Data

We used two kinds of training data in this experiment: Japanese Wikipedia data and "Yahoo Answers Data" from BCCWJ of the National Institute for Japanese Language and Linguistics. For this experiment, the data from Wikipedia were extracted from

jawiki-latest-pages-articles.xml.bz2

which is published on the website for this experiment.

We preprocessed the Wikipedia data before training to obtain Hiragana-only sentences. The preprocessing procedures are as follows. First, we conducted morphological analysis on the Wikipedia data using MeCab. The output of MeCab has features: word segmentation, part of speech, part of speech subdivision 1, part of speech subdivision 2, part of speech subdivision 3, conjugation type, conjugated and basic forms, reading, and pronunciation.

Table 2. MeCab's analysis result for "僕は君と遊ぶ"

|  | 僕<br>*(Boku)* | は<br>*(Ha)* | 君<br>*(Kimi)* | と<br>*(To)* | 遊ぶ<br>*(Asobu)* |
|---|---|---|---|---|---|
| Part of speech | Noun | Particle | Noun | Particle | Verb |
| Part-of-speech subdivision 1 | Pronoun | Binding particle | pronoun | Case particle | Non-bound |
| Part-of-speech subdivision 2 | General | * | General | General | * |
| Part-of-speech subdivision 3 | * | * | * | * | * |
| Conjugation type | * | * | * | * | Godan_verb_ba_column |
| Conjugated form | * | * | * | * | Basic form |
| Basic form | 僕<br>*(Boku)* | は<br>*(Ha)* | 君<br>*(Kimi)* | と<br>*(To)* | 遊ぶ<br>*(Asobu)* |
| Reading | ボク<br>*(Boku)* | ハ<br>*(Ha)* | キミ<br>*(Kimi)* | ト<br>*(To)* | アソブ<br>*(Asobu)* |
| Pronunciation | ボク<br>*(Boku)* | ワ<br>*(Wa)* | キミ<br>*(Kimi)* | ト<br>*(To)* | アソブ<br>*(Asobu)* |

Let us take the example of the sentence, "僕は君と遊ぶ" (*Bokuwakimi to asobu*). Table 2 shows the output result of MeCab when we input this example sentence. This is a Kanji-Kana mixed sentence, which means "I play with you." If this sentence is expressed entirely in Hiragana, that would be "ぼくはきみとあそぶ."

We obtained Hiragana data by replacing the surface forms of words with their readings. Next, we split the Hiragana data into individual characters and assign a part-of-speech tag to each of them.

Here, B-{Part-of-Speech} is assigned to the first Hiragana character, and I-{Part-of-Speech} is assigned to the following Hiragana characters, if the Kanji consisted of more than two syllables.

By formatting this Hiragana sentence as described above, we can obtain the character data and tags corresponding to the character data with one-to-one correspondence (Table 3).

Table 3. Example of splitting "ぼくはきみとあそぶ"

| ぼ (bo) | く (ku) | は (ha) | き (ki) | み (mi) | と (to) | あ (a) | そ (so) | ぶ (bu) |
|---|---|---|---|---|---|---|---|---|
| B-Noun | I-Noun | B-Particle | B-Noun | I-Noun | B-Particle | B-Verb | I-Verb | I-Verb |

Please note that Japanese Kanji characters often have more than one reading. For example, "君" could be Kimi or Kun in Japanese. Therefore, sometimes this conversion makes some errors. The number of characters in the training data is 1,183,624.

We used the original Japanese Wikipedia data for the model that uses Kanji-Kana mixed sentences. Table 4 shows the characters and their tags of the Kanji-Kana mixed sentence, "僕は君と遊ぶ."

Table 4. Example of splitting "僕は君と遊ぶ"

| 僕 (boku) | は (ha) | 君 (kimi) | と (to) | 遊ぶ (aso) | ぶ (bo) |
|---|---|---|---|---|---|
| B-Noun | B-Particle | B-Noun | B-Particle | B-Verb | I-Verb |

For Yahoo! Answers data, we extracted the reading data from the corpus.

Native Japanese speakers manually annotate them, but sometimes they could be different from the authors' intent. For example, "日本" (Japan in Japanese) has two readings, "Nihon" and "Nippon," and both of them are correct in most sentences. Therefore, in these cases, only the author can guarantee which one is correct. In BCCWJ, some words had a predefined reading to reduce the burden of annotators.

### 4.3. Test Data

We also used the BCCWJ for the test data.

The BCCWJ provides sub-corpora and we used 12 of them.

The number of characters used in this experiment for each dataset is shown in Table 5.

Table 5. Number of characters for each sub-corpus in BCCWJ

| Dataset Name | Number of characters |
|---|---|
| Books (Library sub-corpus) | 1,374,216 |
| Bestsellers | 1,093,860 |
| Yahoo! Answers | 830,960 |

| Legal Documents | 2,316,374 |
|---|---|
| National Diet Minutes | 2,050,400 |
| PR Documents | 2,151,126 |
| Textbooks | 956,927 |
| Poems | 466,878 |
| Reports | 2,546,307 |
| Yahoo! Blogs | 1,305,660 |
| Books | 1,281,251 |
| Newspapers | 1,301,728 |

The test data are tagged Hiragana characters with a one-to-one correspondence between the Hiragana character and the tag-based on reading and part-of-speech information, were created in the same way as the training data.

The data from "Yahoo! Answers" are also used as training data for creating the Hiragana Yahoo! model, the Hiragana Wiki + Hiragana Yahoo! model, and the Kanji-Kana Wiki+ Hiragana Wiki+ Hiragana Yahoo! model, however we used different parts for the training and testing.

## 5. RESULT

According to genres of the test data, Tables 6 and 7 summarize the accuracies of the Hiragana Wiki model, Kanji-Kana Wiki+ Hiragana Wiki model, Kanji-Kana Wiki+ Hiragana Wiki+ Hiragana Yahoo! model, Hiragana Yahoo! model, and Hiragana Wiki+ Hiragana Yahoo! model. For the evaluation of each test dataset, we used macro and micro-averages of accuracy. Macro represents the macro-averaged accuracy, and micro represents the micro-averaged accuracy in Tables 6 and 7. In Table 6,blue numbers indicate that the accuracy of the Kanji-Kana Wiki+ Hiragana Wiki model or the Kanji-Kana Wiki+ Hiragana Wiki+ Hiragana Yahoo! model was lower than that of the Hiragana Wiki model, and an asterisk indicates that this difference was significant according to a chi-square test for accuracy at the 5% significance level. In Table 7, Magenta numbers indicate that they are lower than the accuracy of the Kanji-Kana Wiki + Hiragana Wiki+ Hiragana Yahoo! model, and italics indicate that they are lower than the accuracy of the Hiragana Yahoo! model.An asterisk indicates that the difference was significant according to a chi-square test for accuracy at the 5% significance level. A plus indicates that the Hiragana Wiki+ Hiragana Yahoo! model was different from the Hiragana Yahoo! model according to a chi-square test for accuracy at the 5% significance level.The bold numbers are the best results of the models in Tables 6 and 7.

For the Kanji-Kana Wiki+ Hiragana Wiki+ Hiragana Yahoo! model, the Hiragana Yahoo! model, and the Hiragana Wiki+ Hiragana Yahoo! model, the accuracies of Yahoo! Answers, are results of the closed test, and that is why they are written in parentheses.Also, the Yahoo! Answers evaluation data are removed from the macro and micro averages of all models. Therefore, they are average of 11 sub-corpora except for Yahoo! Answers data.

Table 6. Accuracy of Hiragana Wiki model, Kanji-Kana Wiki+ Hiragana Wiki model, and Kanji-Kana Wiki+ Hiragana Wiki+ Hiragana Yahoo! model according to each text genre.

|  | Hiragana Wiki | Kanji-Kana Wiki+ Hiragana Wiki | Kanji-Kana Wiki+ Hiragana Wiki+ Hiragana Yahoo! |
|---|---|---|---|
| Books (Library sub-corpus) | 56.98 | *57.15 | *62.76 |
| Bestsellers | 50.14 | *50.71 | *60.48 |
| Yahoo! Answers | 48.32 | *50.23 | (*65.50) |
| Legal Documents | **63.39** | 63.32 | *60.55 |
| National Diet Minutes | 48.49 | *49.59 | *60.63 |
| PR Documents | 64.27 | *63.93 | *64.55 |
| Textbooks | 57.43 | *58.01 | *62.04 |
| Poems | 41.96 | *42.45 | *48.82 |
| Reports | 65.83 | *65.57 | *65.15 |
| Yahoo! Blogs | 57.12 | 57.10 | *62.97 |
| Books | 55.73 | 55.77 | *62.83 |
| Newspapers | 62.58 | *62.30 | *64.50 |
| Macro | 56.72 | 56.90 | 61.39 |
| Micro | 58.67 | 58.80 | 62.44 |

Blue numbers mean they are lower than the accuracy of the Hiragana Wiki model and an asterisk indicates that the model was different from the Hiragana Wiki model according to a chi-square test at the 5% level of significance. The bold numbers are the best results of the models.

Table 7. Accuracy of Kanji-Kana Wiki+ Hiragana Wiki+ Hiragana Yahoo! model, Hiragana Yahoo! model, and Hiragana Wiki+ Hiragana Yahoo! model according to each text genre.

| | Kanji-Kana Wiki+ Hiragana Wiki+ Hiragana Yahoo! | Hiragana Yahoo! | Hiragana Wiki+ Hiragana Yahoo! |
|---|---|---|---|
| Books (Library sub-corpus) | 62.76 | *63.11 | ***+63.52** |
| Bestsellers | 60.48 | *60.16 | ***+61.32** |
| Yahoo! Answers | (65.50) | (*65.85) | ***+66.38** |
| Legal Documents | 60.55 | *60.17 | *+62.63 |
| National Diet Minutes | 60.63 | *60.49 | ***+61.97** |
| PR Documents | 64.55 | **\*65.49** | *+*63.91* |
| Textbooks | 62.04 | 61.97 | ***+62.66** |
| Poems | 48.82 | *47.61 | ***+49.55** |
| Reports | 65.15 | **\*66.89** | *+*64.89* |
| Yahoo! Blogs | 62.97 | **\*64.02** | +*62.90* |
| Books | 62.83 | *63.28 | ***63.36** |
| Newspapers | 64.50 | **\*65.54** | +64.51 |
| Macro | 61.39 | 61.70 | 61.93 |
| Micro | 62.44 | 62.93 | 63.01 |

Magenta numbers indicate that they are lower than the accuracy ofthe Kanji-Kana Wiki + Hiragana Wiki+ Hiragana Yahoo! model, and italics indicate that they are lower than the accuracy of the Hiragana Yahoo! model. An asterisk indicates that the model was different from the Kanji-Kana Wiki + Hiragana Wiki+ Hiragana Yahoo! model according to a chi-square test at the 5% significance level. A plus indicates that the model is different from Hiragana Yahoo! model at the 5% significance level. The bold numbers are the best results of the models.

We also evaluated the training data using MeCab. For the dictionary of MeCab, we used ipadic with the conversion of Kanji into Hiragana. The macro-averaged accuracy was 79.71%, and the micro-averaged accuracy was 80.10%.Please note that this result cannot simply be compared with results in Tables 6 and 7 because our system did not use a dictionary for the morphological analysis itself.

## 6. DISCUSSION

Table 6 shows that the Kanji-Kana Wiki+ Hiragana Wiki model's macro and micro-averaged accuracies (56.90% and 58.80%) are higher than those of the Hiragana Wiki model (56.72% and 58.67%). The macro-averaged accuracy of the Kanji-Kana Wiki+ Hiragana Wiki model improved by 0.18 points, while the micro-averaged accuracy by 0.13 points. This result indicates that the fine-tuning using the Kanji-Kana Wiki model is somehow effective.

Furthermore, the macro and micro-averaged accuracies of the Kanji-Kana Wiki+ Hiragana Wiki+ Hiragana Yahoo! model (61.39% and 62.44%) are superior to those of the Kanji-Kana Wiki+ Hiragana Wiki model (56.90% and 58.80%). The Kanji-Kana Wiki+ Hiragana Wiki+ Hiragana Yahoo! model improved the accuracy by 4.49 points on the macro-averaged accuracy and by 3.64 points on the micro-averaged, indicating that further fine-tuning using Hiragana Yahoo! data considerably improve the permanence. Additionally, from Tables 6 and 7, we can confirm that the macro- and micro-averaged accuracies of the Kanji-Kana Wiki+ Hiragana Wiki+ Hiragana Yahoo!, Hiragana Yahoo!, and Hiragana Wiki+ Hiragana Yahoo! models, the models using Yahoo! Answers as training data, are higher than those of Wikipedia and Kanji-Kana Wiki+ Hiragana Wiki models, the models using only Wikipedia as training data. In other words, the performance of the model is better when using Yahoo! Answers. We believe there could be two reasons for this result, the quality of the corpus and the similarity of the training and test data. The Hiragana Yahoo! data quality could be better than the Hiragana Wiki data because Hiragana Yahoo! data are manually annotated, whereas Hiragana Wiki data are automatically generated. Moreover, because the training and test data are both sub-corpora of BCCWJ, they can be more similar than when the training and test data are Wikipedia and BCCWJ.

Additionally, comparing the macro- and micro-averaged accuracies of the Kanji-Kana Wiki+ Hiragana Wiki+ Hiragana Yahoo! model with those of the Hiragana Yahoo! model and the Hiragana Wiki+ Hiragana Yahoo! model in Table 7, we confirmed that the Hiragana Wiki+ Hiragana Yahoo! model is the best, and the Hiragana Yahoo! model is the second best, and the Kanji-Kana Wiki+ Hiragana Wiki+ Hiragana Yahoo! model is the last. This result indicates that when the Hiragana Yahoo! data are available, the fine-tuning using both the Kanji-Kana Wiki model and the Hiragana Wiki model is not effective, although the fine-tuning using only the Hiragana Wiki model is useful. Notably, the accuracy of some types of test data is improved while others are decreased. The Kanji-Kana Wiki+ Hiragana Wiki+ Hiragana Yahoo! model was higher than the Hiragana Yahoo! model for the "Bestsellers," "Legal Documents," "National Diet Minutes," "Textbook," and "Poems", and it was higher than the Hiragana Wiki+ Hiragana Yahoo! model for "PR Documents", "Reports", and "Yahoo! Blogs" but there was no genre where the Kanji-Kana Wiki+ Hiragana Wiki+ Hiragana Yahoo! model was the best in these three models.Therefore, we think that the reason why the Kanji-Kana Wiki model was useful when only the Hiragana Wiki data were available could be that the Hiragana Wiki data were automatically generated.

Now, let us discuss the difference among the genres of the texts. The genres where the accuracy of the Hiragana Wiki model was more than 60% were four genres: "Legal Documents," "PR Documents," "Reports," and "Newspapers."

We believe this is because the writing style of these test data is close to that of Wikipedia. Therefore, we marked these genres with underline in Tables 6 and 7. Additionally, we can see that, as for these genres, the fine-tuning is not useful. The accuracies of the Kanji-Kana Wiki+ Hiragana Wiki model decreased for these four genres, and those of the Kanji-Kana Wiki+ Hiragana Wiki+ Hiragana Yahoo! model also decreased for two genres. As for the remaining two genres, the improvements are less than two points.

Figure 2 shows the effects or the differences of the accuracies of Hiragana Yahoo! data according to the genres of the texts. The blue line is its effects when the base model is the Kanji-Kana Wiki+ Hiragana Wiki model and the orange line is when the base model is the Hiragana Wiki model. This figure shows that the effects are the same even though the base models are different.

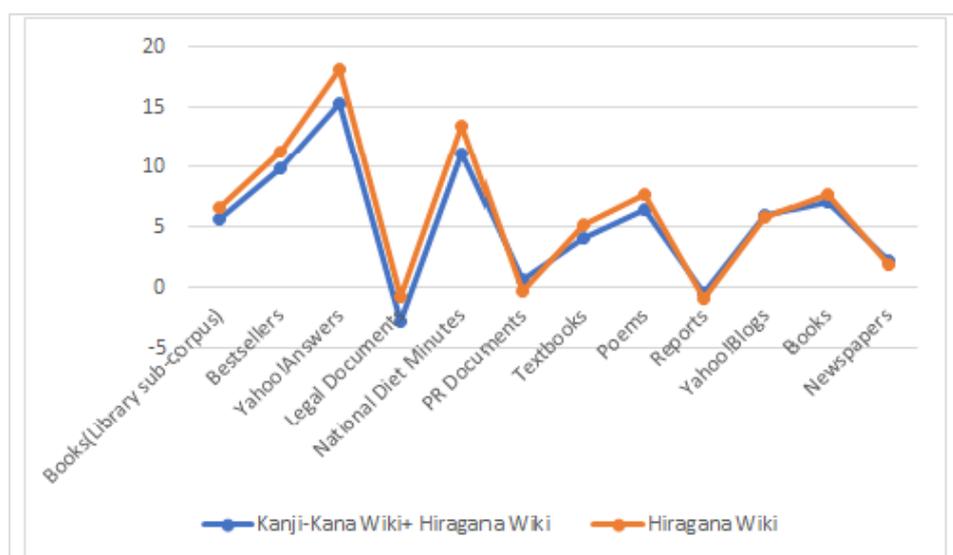

Figure 2. Effect of Hiragana Yahoo! data

Hiragana Yahoo! data was considerably effective for "Yahoo! Answers," "Books," and "National Diet Minutes." Hiragana Yahoo! data is text data of the question answering sites. Therefore, the writing style is rather like spoken language. We believe that it is effective for "Books" and "National Diet Minutes" because the Hiragana Books include dialogues and National Diet Minutes is transcription of the discussion of the Diet. In particular, Wikipedia data rarely include question sentences but question answering sites contain many. Therefore, Hiragana Yahoo! data is effective for "National Diet Minutes" that includes many questions.

Figure 3 shows the effects or the differences of the accuracies of Kanji-Kana Wiki data according to the genres of the texts. The blue line is its effects when the base model is the Hiragana Wiki model, and the orange line is when the base model is the Hiragana Wiki+ Hiragana Yahoo! model. As contrasted to Figure 2, the effects of the data are almost opposite depending on the base model. The blue line is similar to the lines in Figure 2 but the orange line is very different from those. The orange line, the effects of the Hiragana Yahoo! data when it was added to the Hiragana Wiki+ Hiragana Yahoo! model shows that the data is effective for "PR Documents,""Reports," and"Yahoo! Blogs." The Kanji-Kana Wiki data tends to effective when the fine-tuning of the Hiragana Yahoo! is not effective. These facts indicate that the fine-tuning is effective when the original accuracy was poor, regardless of whether it used Kanji-Kana Wiki text or Hiragana Yahoo! data.

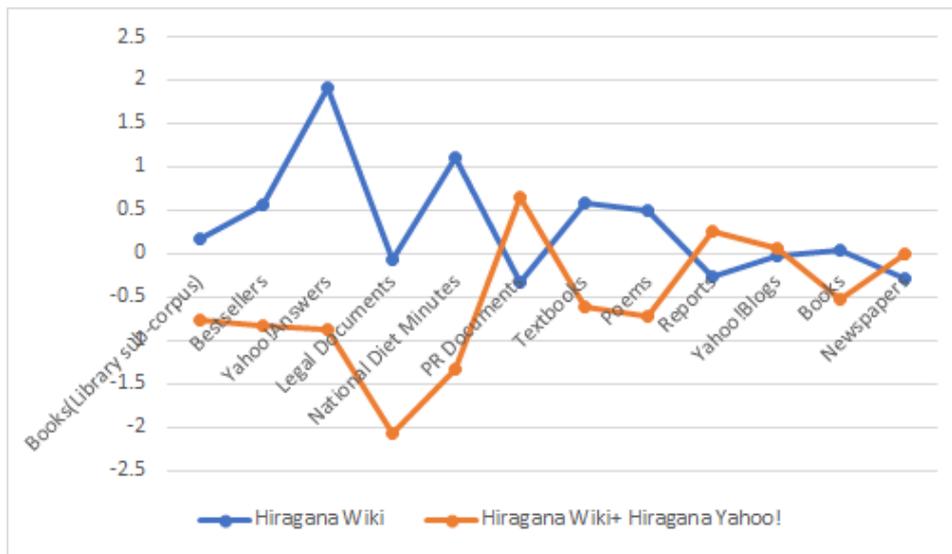

Figure 3. Effect of Kanji-Kana Wiki data

Finally, the accuracy of the Bi-LSTM CRF model still has room for improvement compared to the existing morphological analyzer for Kanji-Kana mixed text. We think that the use of dictionary and much more data should be attempted in the future.

## 7. CONCLUSIONS

This study developed a morphological analysis model for Japanese Hiragana sentences using the Bi-LSTM CRF model.We showed that the performance of morphological analysis of Hiragana sentences outperformed when we use in-domain data manually annotated for fine-tuning by the experiments using Hiragana data of Wikipedia and Yahoo! Answers data. We also showed that the performance of Hiragana morphological analysis is improved when we fine-tune the model of Hiragana sentences with Kanji-Kana mixed sentences. This fine-tuning is effective when the original accuracy was low. Additionally, we showed that the performance of morphological analysis varied according to the text genre. In the future, we would like to increase the number of training data to improve the accuracy of the model for analyzing Hiragana sentences. Additionally, we would like to make the system capable of outputting word reading, pronunciation, and part-of-speech classification.In this experiment, we used the Yahoo! Answers as a training example for the Kanji-Kana Wiki+ Hiragana Wiki+ Hiragana Yahoo! model and Hiragana Yahoo! models to measure the accuracy, but we would like to train the models using other data to analyze the effect of genre differences.Furthermore, we would like to investigate the effective domain adaptation method when the original accuracy is not bad.

## REFERENCES

[1] Kudo, Taku., Yamamoto, K., and Matsumoto, Y. (2004). "Applying conditional random fieldsto Japanese morphological analysis." *In Proceedings of the 2004 Conference on Empirical Methods in Natural Language Processing*, pp. 230–237.
[2] Morita, Hajime., Kawahara, D., and Kurohashi, S. (2015). "Morphological analysis for unsegmented languages using recurrent neural network language model." *In Proceedings of the 2015 Conference on Empirical Methods in Natural Language Processing*, pp. 2292–2297.
[3] Tolmachev, Arseny, Daisuke Kawahara, and Sadao Kurohashi. "Juman++: A morphological analysis toolkit for scriptio continua." *Proceedings of the 2018 Conference on Empirical Methods in Natural Language Processing*. 2018, pages 54-59.

[4] Kikuo Maekawa, Makoto Yamazaki, Toshinobu Ogiso,Takehiko Maruyama, Hideki Ogura, Wakako Kashino, Hanae Koiso, Masaya Yamaguchi, Makiro Tanaka, and Yasuharu Den. 2014. "Balanced Corpus of Contemporary Written Japanese".*Language resources and evaluation*, 48(2):345–371.

[5] Jun Izutsu, Riku Akashi, Ryo Kato, Mika Kishino, Taichiro Kobayashi, Yuta Konno, and Kanako Komiya. 2020. "Morphological analysis of hiragana-only sentences using MeCab". Inthe Proceedings of*NLP2020*, pages 65–68 (In Japanese).

[6] Taku Kudo, Hiroshi Ichikawa, David Talbot, and Hideto Kazawa. 2012. "Robust morphological analysis for hiragana sentences on the web". In the Proceedings of NLP 2012, pages 1272–1275 (In Japanese).

[7] Ayaha Osaki, Syouhei Karaguchi, Takuya Ohaku, Syunya Sasaki, Yoshiaki Kitagawa, Yuya Sakaizawa, and Mamoru Komachi. 2016. "Corpus construction for Japanese morphological analysis in twitter". In the *Proceedings of NLP 2016*, pages 16–19 (In Japanese).

[8] Sanae Fujita, Hirotoshi Taira Tessei Kobayashi, and Takaaki Tanaka. 2014. "Morphological analysis of picture book texts".*Journal of Natural Language Processing*, 21(3):515–539 (In Japanese).

[9] Masato Hayashi and Takashi Yamamura. 2017. "Considerations for the addition of hiragana words and the accuracy of morphological analysis". In *Thesis Abstract of School of Information Science and Technology*, Aich Prifectual University, pages 1–1 (In Japanese).

[10] Ji Ma, Kuzman Ganchev, and David Weiss. 2018. "State-of-the-art Chinese word segmentation with Bi-LSTM".In Proceedings of *the 2018 Conference on Empirical Methods in Natural Language Processing*, pages4902–4908. Association for Computational Linguistics.

[11] Suphanut Thattinaphanich and Santitham Prom-on.2019. "Thai named entity recognition usingBi-LSTM-CRF with word and character representation". In *2019 4th International Conference on Information Technology (InCIT)*, pages 149–154.

[12] Tolmachev, Arseny, Daisuke Kawahara, and Sadao Kurohashi. "Shrinking Japanese morphological analyzers with neural networks and semi-supervised learning." *Proceedings of the 2019 Conference of the North American Chapter of the Association for Computational Linguistics: Human Language Technologies*, Volume 1 (Long and Short Papers). 2019, pages 2744–2755

[13] Chen, Xinchi, Xipeng Qiu, Chenxi Zhu, Pengfei Liu, and Xuanjing Huang. "Long short-term memory neural Networks for Chineseword segmentation." *Proceedings of the 2015 Conference on Empirical Methods in Natural Language Processing*. 2015, pages 1197-1206

[14] Kikuo Maekawa, Makoto Yamazaki, Takehiko Maruyama, Masaya Yamaguchi, Hideki Ogura, Wakako Kashino, Toshinobu Ogiso, HanaeKoiso, and Yasuharu Den. 2010. "Design, compilation, and preliminary analyses of Balanced Corpus ofContemporary Written Japanese". In Proceedings of *the Seventh International Conference on Language Resources and Evaluation* (LREC 2010), pages1483–1486.